\documentclass[runningheads]{llncs}

\usepackage{eccv}

\usepackage{eccvabbrv}

\usepackage{graphicx}
\usepackage{booktabs}
\usepackage{pifont}
\usepackage[accsupp]{axessibility}  

\usepackage{hyperref}

\usepackage{orcidlink}
\usepackage{multicol}
\usepackage{multirow}
\newcommand{\mymodel}{CausalDrive}

\usepackage{pifont}
\usepackage{wasysym}

\newcommand{\corrauthor}{\textsuperscript{\ding{171}}}
\newcommand{\projectlead}{\textsuperscript{\textdagger}}
\newcommand{\intern}{\textsuperscript{$\ddagger$}} 

\begin{document}

\title{\mymodel: Real-time Causal World Models for Autonomous Driving} 

\titlerunning{Abbreviated paper title}

    \author{Tianyi Yan\intern\inst{1} \and
Huan Zheng\inst{1} \and
Dubing Chen\inst{1} \and
Meizhi Qu\inst{3} \and
 Yingying Shen\inst{2} \and
 Lijun Zhou\inst{2} \and
Mingfei Tu\inst{2} \and
Bing Wang\inst{2} \and 
Guang Chen\inst{2}\and
Hangjun Ye\inst{2}\and 
Haiyang Sun\projectlead\inst{2} \and
Cheng-zhong Xu\inst{1} \and
Jianbing Shen\corrauthor\inst{1}\thanks{\ding{171} Corresponding author. \textdagger Project Lead. $\ddagger$ Intern in Xiaomi EV.}
}

\authorrunning{F.~Author et al.}

\institute{SKL-IOTSC, CIS, University of Macau \and
Xiaomi EV \and
CASIA
}

\maketitle

\begin{abstract}
  World models have emerged as a promising paradigm for scaling autonomous driving (AD) data, yet existing video generative models fall short as interactive simulators. Layout-conditioned renderers rely on "oracle" future trajectories of all background agents, rendering them strictly non-reactive. Conversely, pure action-conditioned predictors lack semantic control over complex interactions and suffer from prohibitive diffusion latencies, hindering closed-loop policy learning. 
  To bridge this gap, we present CausalDrive, a controllable, real-time foundation driving world renderer. CausalDrive operates solely on the initial front-view frame, the ego-vehicle’s trajectory, and a macroscopic text prompt. By excluding future NPC layouts, we compel the model to intrinsically predict causal interactions, enabling text-driven control over Driving Sociology—allowing users to dynamically orchestrate diverse counterfactual reactions to identical ego-actions. To overcome the efficiency bottleneck and address the covariate shift in autoregressive generation, we propose a novel Context-Forced DMD architecture. This combines continuous flow-matching with a self-correcting distillation objective, achieving interactive speeds of 12 FPS. This breakthrough transforms the passive video generator into a playable neural simulator. We demonstrate its versatility across three downstream applications: (1) generative closed-loop evaluation with significantly mitigated collision artifacts, (2) large-scale Reinforcement Learning (RL) post-training driven by a Video2Reward module, and (3) real-time human-in-the-loop simulation. Extensive experiments validate that policies trained within CausalDrive’s reactive scenarios exhibit superior interaction capabilities in the real world.
\end{abstract}



\section{Introduction}
\label{sec:intro}
The paradigm of Autonomous Driving (AD) is shifting from modular pipelines to end-to-end learning~\cite{hu2023uniad,jiang2023vad,chen2024vadv2,li2025recogdrive,liao2025diffusiondrive}. 
Within this transition, learning a world model~\cite{yan2024drivingsphere,wang2024drivedreamer,zhao2025drivedreamer2,gao2023magicdrive,gao2024magicdrivedit,wen2024panacea,yan2025olidm} capable of simulating plausible future outcomes is envisioned as a key milestone toward high-level machine intelligence. 
Recent advances~\cite{ho2020denoising,blattmann2023svd} in generative AI have achieved remarkable success in high-fidelity visual synthesis, precise instruction controllability~\cite{gao2023magicdrive,gao2024magicdrivedit,liang2022bevfusion}, and the simulation of safety-critical corner cases~\cite{tian2025simscale,yan2025ad}. 

However, a critical dichotomy remains. Existing models~\cite{gao2023magicdrive,gao2024magicdrivedit,yan2025olidm,liang2025worldlens} primarily function as "offline data engines" for augmentation, rather than interactive environments. 
SOTA driving world models~\cite{gao2023magicdrive,liang2022bevfusion,gao2024magicdrivedit,wen2024panacea,li2024uniscene} exhibit stunning photorealism but largely operate as "open-loop" video predictors.
Relying heavily on log-replay, these models often ignore the ego-vehicle’s novel actions or fail to model the causal reactions of surrounding agents (e.g., a neighbor yielding when the ego merges). 
Consequently, the generated environment remains a passive "background" rather than an active participant. 
Furthermore, the prohibitive inference latency of diffusion-based architectures (often seconds per frame) renders them impractical for computationally intensive tasks like online Reinforcement Learning (RL).

To serve as a viable alternative to real-world data collection and casual-aware simulators, we argue that the next generation of models must evolve into Foundation Driving World Renderers, holistic systems that are not just "dreamers" of video, but real-time, reactive, and physically grounded.
Inspired by recent discussions on driving simulation~\cite{yang2024drivearena,yan2024drivingsphere,yang2025resim}, we identify four essential pillars required to unlock the full potential of world models for downstream applications:
(1)High-Fidelity Visual Synthesis (Photorealism): To minimize the domain gap, the model must produce sensor-realistic outputs where perception networks can generalize directly from simulation to reality. 
(2) Strict Action-Controllability: The generated dynamics must faithfully reflect control signals (steering, throttle) to prevent "action-ignoring" or posterior collapse, ensuring a "left turn" command results in a physically accurate visual transition. 
(3) Data-Driven Reactive Agents (Driving Sociology): Unlike rigid "log-replay" systems, the model must implicitly capture the "sociology" of driving, the causal, reactive interactions between the ego-vehicle and surrounding agents (e.g., yielding, merging, or reacting to hazards) 
(4) Real-Time Inference Efficiency: Simulation-based applications, such as large-scale reinforcement learning (RL) or human-in-the-loop testing, require interactive frame rates ($>$ 10 FPS) to be computationally feasible. 

Despite individual advancements~\cite{gao2024vista,yang2025resim,li2024uniscene}, integrating these four pillars into a unified framework remains an open challenge. Current models often trade latency for quality or sacrifice interactivity for stability. In this work, we propose \textbf{\mymodel}, a Foundation Driving World Renderer designed to satisfy these desiderata simultaneously. 

To systematically address the challenges of driving sociology and real-time inference, we introduce \textbf{CausalDrive}, an Interactive Driving World Model designed to unify photo-realistic generation with closed-loop reactive logic. Our framework encompasses three core innovations across data curation, model architecture, and acceleration.
To cultivate predictive driving sociology without relying on "oracle" layout leaks, we establish SocioDrive-Bench. Unlike layout-conditioned models that passively render future bounding boxes, CausalDrive predicts futures strictly conditioned on the initial frame, ego-trajectory, and semantic prompts. By leveraging a vision-language pipeline to extract explicit causal interaction labels—such as active pressure, defensive driving, and complex negotiation—we transform raw geometric logs into sociologically-aware training curricula.
Architecturally, directly converting a bidirectional video diffusion model into an autoregressive (AR) streaming simulator causes expectation collapse due to the violation of probability flow injectivity. We bridge this gap by constructing a Sociology-Aware Causal AR Teacher. Conditioned via adaptive layer normalization for geometric poses and cross-attention for semantic sociology prompts, this continuous flow-matching teacher ensures mathematically sound and dynamically accurate step-by-step rollouts.

Finally, to overcome exposure bias and achieve real-time simulation speeds, we propose Self-Corrective Forcing (SCF) for diffusion distillation. Autoregressive generation over long horizons inherently accumulates errors. Standard distillation fails in this regime due to a context mismatch between the teacher's perfect history and the student's flawed history. By explicitly forcing the AR teacher to evaluate gradients based on the student's imperfect self-rollout, SCF teaches the student to auto-correct temporal compounding errors. This reduces the required function evaluations to $\le 4$ steps, achieving a throughput of $\ge 12$ FPS with sub-second latency.

By integrating these innovations, CausalDrive effectively transforms a passive video generator into a playable neural simulator. We demonstrate its versatility across three downstream applications: closed-loop evaluation with highly consistent kinematic dynamics, efficient RL training utilizing a Video-to-Reward (V2R) module, and human-in-the-loop simulation with real-time WASD control.

Our main contributions are summarized as follows:
\begin{itemize}
    \item We propose \textbf{CausalDrive}, a real-time foundation world model that strictly avoids future-layout conditioning, unifying photo-realistic generation with closed-loop reactive logic for autonomous driving.
    \item We introduce \textbf{SocioDrive-Bench}, establishing a novel VLM/LLM-driven taxonomy to extract and formalize causal driving sociology (e.g., yielding, negotiating), enabling semantic control over multi-agent interactions.
    \item We identify the architectural and context-mismatch bottlenecks in streaming video generation, and propose a novel \textbf{Context-Forced DMD} framework over a \textbf{Causal AR Flow-Matching Teacher}. This enables robust, infinite-horizon generation at $\ge 12$ FPS.
    \item Extensive experiments demonstrate that CausalDrive acts as a superior neural simulator. RL policies trained safely within its hallucinated, safety-critical scenarios exhibit significantly improved interaction capabilities.
\end{itemize}

\begin{figure}[t]
    \centering
    \includegraphics[width=0.97\linewidth]{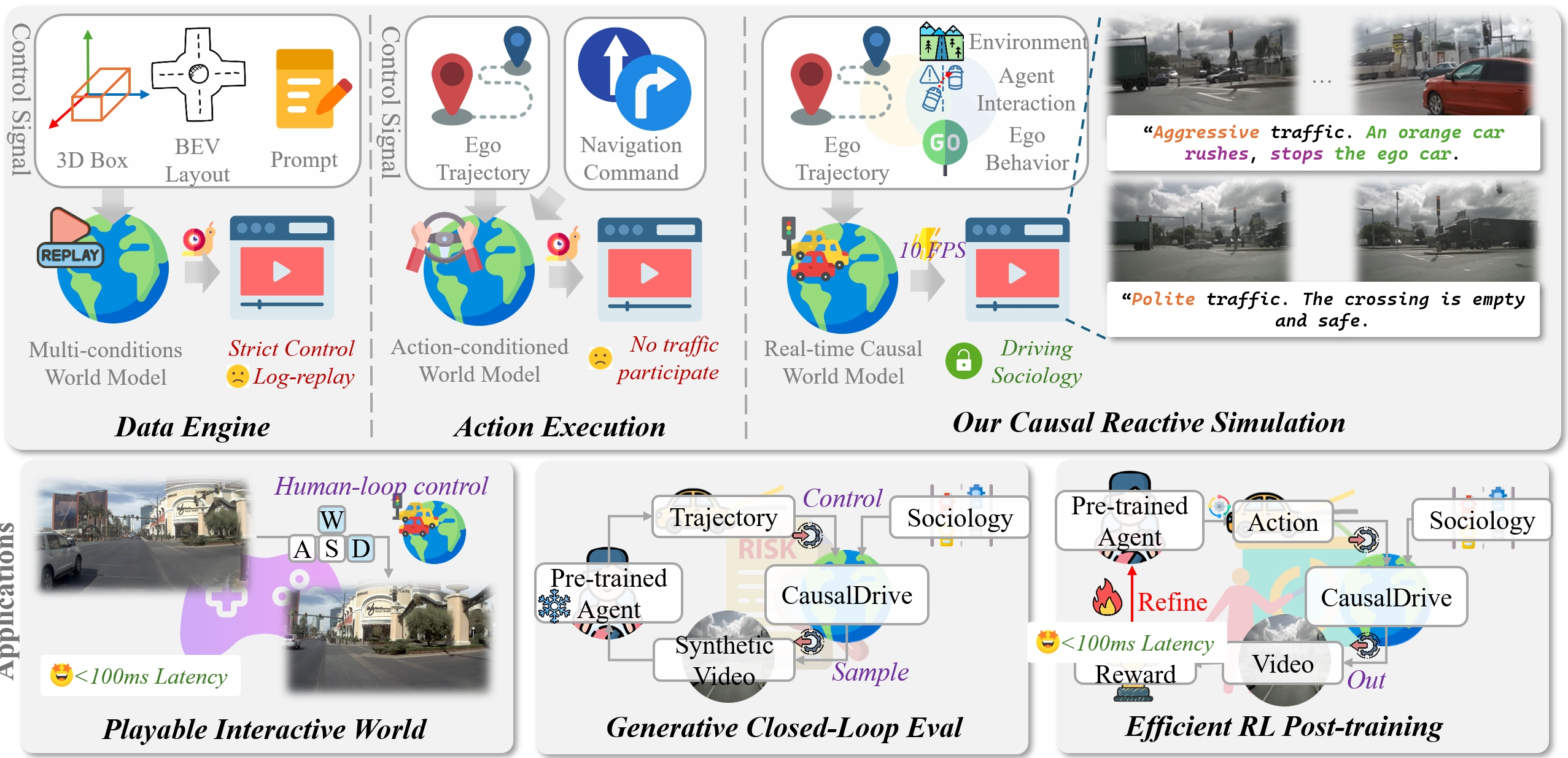}
    \caption{
    \textbf{Top: Semantic Control of Driving Sociology:} Unlike layout-conditioned renderers that require "oracle" future boxes of all agents, \mymodel~generates interactive futures. Given the \textit{exact same} initial state and ego-trajectory (nudging into an intersection), modifying the prompt allows us to dynamically simulate diverse counterfactual reactions from surrounding agents (e.g., politely yielding vs. aggressively rushing). 
    \textbf{Down: The reactive API that powers three distinct downstream applications:} (1) \textbf{Playable Interactive World}, offering a playable, low-latency virtual world controlled via WASD/steering wheels. (2) \textbf{Closed-Loop Evaluation}, providing a "zero-ghosting" proving ground for planners; and (3) \textbf{RL Post-Training}, where an agent safely learns from simulated safety-critical scenarios via a Video2Reward (V2R) module.}
    \label{fig:placeholder}
        \vspace{-2mm}

\end{figure}
\section{Related Work}
\subsection{Generative World Models for Autonomous Driving}
\label{subsec:rw_world_models}

The advent of large-scale generative models has catalyzed the development of world models for autonomous driving (AD), aiming to synthesize infinite driving scenarios for data augmentation and policy learning~\cite{hu2023gaia, wang2024drivedreamer,zhao2025drivedreamer2}. Current video-based driving models broadly fall into two paradigms based on their conditioning strategies. 
The first paradigm, \textit{Layout-Conditioned Renderers} (e.g., MagicDrive~\cite{gao2023magicdrive,gao2024magicdrivedit}, UniScene~\cite{li2024uniscene}, Panacea~\cite{wen2024panacea}), synthesizes high-fidelity videos conditioned on the complete future trajectories or 3D bounding boxes of all traffic participants. While producing stable visuals, these models act as "oracles" rather than simulators; they are fundamentally non-reactive because the future states of surrounding agents are hard-coded inputs rather than dynamically inferred responses. 
The second paradigm, \textit{Action-Conditioned Predictors} (e.g., Vista~\cite{gao2024vista}, Orbis~\cite{mousakhan2025orbis}, Drive-WM~\cite{wang2024drivingdrivewm}), predicts future frames based solely on the ego-vehicle's action and past context. While conceptually closer to true world models, they often suffer from "posterior collapse" (ignoring ego-actions) and lack macroscopic semantic control over the diverse, long-tail behaviors of surrounding agents.
In contrast, \textbf{CausalDrive} takes a hybrid approach: we strictly omit future NPC layouts to compel genuine causal prediction, while introducing a macroscopic sociology prompt to semantically orchestrate the predicted interactions, achieving both strict action-controllability and semantic diversity.

\subsection{Reactive Simulation and Driving Sociology}
\label{subsec:rw_simulation}

Closed-loop policy evaluation and Reinforcement Learning (RL) require environments where non-player characters (NPCs) react logically to the ego-vehicle. Traditional graphics-based simulators (e.g., CARLA~\cite{dosovitskiy2017carla}, MetaDrive~\cite{li2022metadrive}) provide interactivity but suffer from a severe visual sim-to-real gap and utilize rigid, rule-based NPCs that fail to capture human-like stochasticity. To address the visual gap, data-driven approaches often rely on \textit{Log-Replay} from real-world datasets (e.g., nuPlan~\cite{dauner2024navsim,caesar2021nuplan}). However, log-replay is inherently open-loop; if the ego-vehicle deviates from the logged trajectory, surrounding vehicles fail to react, leading to the unrealistic "ghost effect" (false-positive collisions)~\cite{yang2024drivearena, yan2024drivingsphere}. While eliminating these artifacts entirely remains an open challenge, CausalDrive significantly mitigates them by strictly enforcing causal dependencies.
We introduce \textbf{SocioDrive-Bench}, shifting the focus from geometric trajectories to \textit{Driving Sociology}. By leveraging Large Vision-Language Models (VLMs), we explicitly extract causal interaction labels (e.g., yielding, negotiating), bridging the gap between raw perceptual logs and cognitive reactive simulation.

\subsection{Efficient Video Generation and Distillation}
\label{subsec:rw_distillation}

Transforming video diffusion models into real-time interactive simulators remains a formidable challenge due to the high Number of Function Evaluations (NFE) required by iterative solvers. While generalized distillation techniques like Consistency Models~\cite{song2023consistency} and Distribution Matching Distillation (DMD)~\cite{yin2024one} have accelerated text-to-image and short-video generation, applying them to open-ended, streaming driving simulation introduces unique bottlenecks.
State-of-the-art video Diffusion Transformers (DiTs)~\cite{peebles2023scalabledit} rely on bidirectional temporal attention. Directly distilling a bidirectional teacher into an Autoregressive (AR) streaming student violates Probability Flow ODE (PF-ODE) injectivity—where multiple valid futures collapse into blurred predictions~\cite{huang2025selfforcing,zhu2026causalforcing}. Furthermore, long-horizon AR generation suffers from \textit{exposure bias}, where small student errors compound over time.
To systematically resolve this, we propose a mathematically sound pipeline: we first train a \textbf{Causal AR Teacher} via Continuous Flow Matching to guarantee conditional alignment, and then introduce \textbf{Context-Forced DMD}. By forcing the teacher to evaluate gradients on the student's flawed autoregressive rollouts, we explicitly correct exposure bias, successfully compressing high-fidelity reactive generation to $\le 4$ steps ($\ge 12$ FPS).

\section{Methodology}
\label{sec:method}
Our ultimate goal is to construct a real-time, interactive driving world model capable of modeling the causal interactions among various traffic participants (agents). 
Let $\mathbf{x} \in \mathbb{R}^{H \times W \times 3}$ denote a visual observation. Instead of passively predicting video continuations, the model must act as a reactive simulator, approximating the transition dynamics $p_\theta(C_{1:K} \mid C_{0}, \mathcal{A}_{1:K})$, where $C_k$ represents a temporal chunk of latent visual states, and $\mathcal{A}_{1:K}$ denotes a streaming sequence of external control signals.

\begin{figure}[t]
    \centering
    \includegraphics[width=0.97\linewidth]{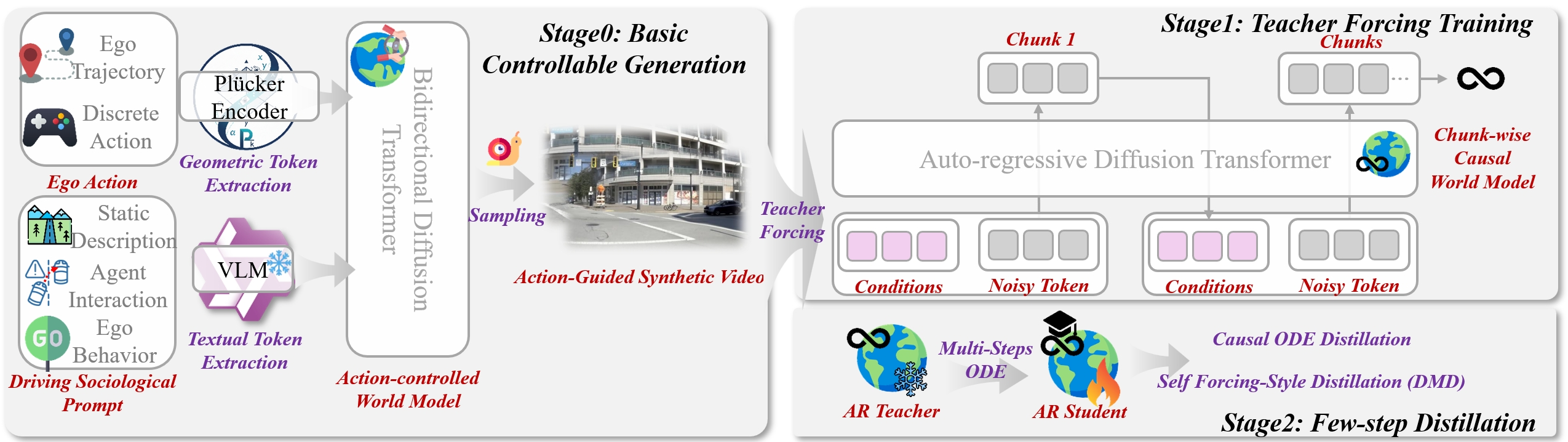}
\caption{\textbf{Architecture and Training Pipeline of CausalDrive.} \textbf{(a) Stage 0:} The base model employs Block Causal Attention for $O(1)$ streaming. Ego-trajectories ($\mathcal{P}$) and sociology prompts ($\mathcal{B}$) are injected via AdaLN and cross-attention, respectively. \textbf{(b) Stage 1:} To prevent probability flow injectivity violations, we train a continuous flow-matching Autoregressive (AR) teacher using strict Teacher Forcing on ground-truth history ($C_{<k}$). \textbf{(c) Stage 2: Self-Corrective DMD.} To mitigate exposure bias during long-horizon rollouts, the student performs an AR self-rollout to generate a flawed memory context ($\hat{C}_{\text{stu}}$). The distillation gradient $\nabla\mathcal{L}_{\text{DMD}}$ is evaluated by forcing the frozen teacher to condition on this flawed context, thereby teaching the student to auto-correct accumulated temporal errors.}
    \vspace{-2mm}
    \label{fig:placeholder}
\end{figure}
Transforming a powerful but slow bidirectional video generative model into an ultra-fast autoregressive (AR) interactive simulator presents two fundamental challenges: (1) the \textit{architectural gap}, where distilling a bidirectional teacher into a causal student violates PF-ODE injectivity, causing blurry generation; and (2) the \textit{context mismatch}, where error accumulates over long-horizon autoregressive rollouts (exposure bias). To systematically address these issues, we formulate our method into three parts. We first provide the preliminaries in Sec.~\ref{sec:preliminary}. Then, we introduce the construction of a driving sociology-aware causal autoregressive teacher in Sec.~\ref{sec:ar_teacher}. Finally, we detail the few-step distillation framework via Causal ODE and Context-Forced DMD in Sec.~\ref{sec:distillation}.
\subsection{Preliminaries}
\label{sec:preliminary}

\textbf{Bidirectional Video Diffusion Models.} 
Recent large-scale video generation heavily relies on continuous-time diffusion models parameterized by Diffusion Transformers (DiTs). 
Let $\mathbf{z}_0 \sim p_{\text{data}}$ denote a real video chunk in the latent space, and $\mathbf{z}_1 \sim \mathcal{N}(\mathbf{0}, \mathbf{I})$ denote standard Gaussian noise. Under the Flow Matching (FM) framework~\cite{lipman2022flow}, the PF-ODE governing the marginal distributions is defined as $\mathrm{d}\mathbf{z}_t = v_\theta(\mathbf{z}_t, t) \mathrm{d}t$, for $t \in [0, 1]$. The optimal transport path corresponds to linear interpolation: $\psi_t(\mathbf{z}_0) = t\mathbf{z}_1 + (1-t)\mathbf{z}_0$.
The network $v_\theta$ is trained to regress the velocity field:
\begin{equation}
    \mathcal{L}_{\text{FM}}(\theta) = \mathbb{E}_{\mathbf{z}_0, \mathbf{z}_1, t} \left[ \left\| v_\theta(\psi_t(\mathbf{z}_0), t) - (\mathbf{z}_1 - \mathbf{z}_0) \right\|^2 \right]
\end{equation}
Standard video DiTs employ $v_\theta$ employs \textit{Bidirectional Attention} across the temporal dimension. When denoising frame $x_t^{(i)}$, the model attends to all other frames, including the future. While excelling at fixed-length generation, this non-causal nature prevents real-time, open-ended interactivity.

\textbf{Chunk-wise Autoregressive Generation.} 
To enable streaming generation, we factorize the joint distribution of a video into non-overlapping temporal chunks $C_1, C_2, \dots, C_K$:
\begin{equation}
    p_\theta(C_{1:K}) = \prod_{k=1}^K p_\theta(C_k \mid C_{<k}, \mathcal{A}_{\le k})
\end{equation}
We adopt a \textit{Block Causal Attention} mask: tokens within the same chunk $C_k$ attend to each other bidirectionally to preserve local spatio-temporal dynamics, while attention to past chunks $C_{<k}$ is strictly causal and cached via a Key-Value (KV) mechanism to ensure $O(1)$ streaming complexity.

\subsection{Driving Sociology-Aware Causal Autoregressive Teacher}
\label{sec:ar_teacher}

Directly distilling a bidirectional model into an AR student causes conditional expectation collapse[casual forcing]. Therefore, we first construct a mathematically sound \textit{Causal Autoregressive Teacher}, denoted as $v_{\theta^*}^{\text{AR}}$.

Crucially, driving is not merely following geometric trajectories; it is a highly interactive multi-agent social process. Thus, we condition our AR teacher on a hybrid control signal $\mathcal{A} = (\mathcal{P}, \mathcal{B})$, establishing a causal hierarchy between low-level ego-mechanics and high-level sociology.

Given the ego-vehicle trajectory $\mathcal{P}$, we encode the continuous camera poses into Plücker coordinates and process them via a Multi-Layer Perceptron (MLP) to obtain spatial modulation parameters $\mathbf{h}_{\text{geo}}$. Simultaneously, the discrete behavioral description $\mathcal{B}$ (e.g., \textit{"ego vehicle yields to an aggressive pedestrian"}) is mapped into continuous semantic embeddings $\mathbf{E}_{\text{socio}}$ via a pre-trained text encoder. It is important to note that $\mathcal{B}$
 serves as a global "sociological prior" (e.g., a "Polite" scene implies a higher probability of yielding across all agents) rather than a rigid script for specific instance-level control. 

Within each DiT block, we explicitly decouple these conditions to model causal interactions. Let $\mathbf{z}$ denote the intermediate latent feature. The geometric action imposes a hard spatial constraint via Adaptive Layer Normalization (AdaLN), while the sociology prompt modulates the multi-agent reaction via Cross-Attention:
\begin{align}
    \hat{\mathbf{z}} &= \mathbf{z} + \text{SelfAttn}\big(\text{AdaLN}(\mathbf{z}, \mathbf{h}_{\text{geo}})\big) \label{eq:adaln} \\
    \mathbf{z}' &= \hat{\mathbf{z}} + \text{CrossAttn}\big(\text{Q}=\hat{\mathbf{z}}, \text{K,V}=\mathbf{E}_{\text{socio}}\big) \label{eq:crossattn}
\end{align}
This architectural design strictly enforces causality: Eq.~\ref{eq:adaln} guarantees that the visual perspective precisely shifts according to the ego-vehicle's steering and throttle, preventing "action-ignoring." Concurrently, Eq.~\ref{eq:crossattn} allows the semantic prior $\mathbf{E}_{\text{socio}}$ to query the spatially-anchored features, hallucinating logical reactions from surrounding agents (e.g., synthesizing a braking neighbor) without interfering with the ego-trajectory's geometric fidelity.

We train this causal AR teacher $v_\theta^{\text{AR}}$ using \textbf{Teacher Forcing}. Specifically, when training the $k$-th chunk, the attention mechanism is strictly conditioned on the \textit{clean, ground-truth} historical prefix $C_{<k}$, rather than a noisy prefix. This guarantees that the training objective perfectly aligns with the inference distribution, establishing a mathematically sound and dynamically accurate AR teacher.

\subsection{Few-Step Distillation}
\label{sec:distillation}

While the AR teacher achieves high fidelity, it requires numerous ODE solver steps (e.g., 50 steps), violating the latency constraints of a real-time interactive world model. We compress the generation to 1-4 steps via a two-stage distillation process.

\subsubsection{Causal ODE Distillation.} 
A fundamental requirement for ODE distillation is \textit{frame-level injectivity}: each noisy frame must map to a unique clean frame. Distilling an AR student from a bidirectional teacher violates this, as the absence of future frames causes one noisy state to map to multiple possible clean outcomes, resulting in blurred predictions. 
To bridge this architectural gap, we perform distillation strictly using trajectories generated by our Causal AR Teacher $v_\theta^{\text{AR}}$. Given a clean history $C_{<k}$ and actions $\mathcal{A}_{\le k}$, we sample exact PF-ODE trajectories from noise to data. The AR student $G_\phi$ is trained to regress the clean target $C_k^{(0)}$ from intermediate noisy states $C_k^{(t)}$:
\begin{equation}
    \phi^* = \arg\min_\phi \mathbb{E}_{C_{<k}, t} \left[ \left\| G_\phi(C_k^{(t)}, C_{<k}, \mathcal{A}_{\le k}, t) - C_k^{(0)} \right\|^2 \right]
\end{equation}
Since both the teacher and student are causal, frame-level injectivity perfectly holds, allowing the student to correctly learn the flow map without detail degradation.

\subsubsection{Context-Forced DMD for Infinite Horizon.} 
To further reduce sampling to 1-4 steps, we apply Distribution Matching Distillation (DMD). However, in autoregressive long-horizon generation, the student accumulates errors and deviates from the training distribution (\textit{Exposure Bias}). Standard DMD fails here because it creates a \textit{Context Mismatch}: the teacher evaluates gradients based on perfect ground-truth history, while the student generates based on its own flawed history. This phenomenon is akin to the "Covariate Shift" problem in imitation learning.

To resolve this, we propose \textbf{Context Forcing}. During training, the student performs an $N$-chunk \textit{Self-Rollout} to construct a flawed memory context $\hat{C}_{stu}$. When computing the real score using the AR Teacher, we explicitly force the Teacher to condition on this student-generated history $\hat{C}_{stu}$, rather than the ground truth:
\begin{equation}
    \nabla_\phi \mathcal{L}_{\text{DMD}} \approx \mathbb{E} \left[ s_{\text{real}}\big(\hat{C}_k^{(t)}, t \mid \hat{C}_{stu}, \mathcal{A}\big) - s_{\text{fake}}\big(\hat{C}_k^{(t)}, t \mid \hat{C}_{stu}, \mathcal{A}\big) \right] \frac{\partial \hat{C}}{\partial \phi}
\end{equation}
By forcing the teacher to ``clean up the student's mess,'' the distillation gradient explicitly teaches the student how to recover from its own errors. Furthermore, to preserve high-frequency road textures (e.g., lane markings) over long simulations, we attach an adversarial discriminator to the fake score network, pushing the few-step generated driving chunks to be indistinguishable from real traffic videos.

\subsection{Downstream Applications Formulation}
\label{subsec:applications_formulation}

By satisfying real-time ($\ge$ 12 FPS) and reactive requirements, \mymodel~naturally serves as a universal interactive API. We formulate three specific downstream integrations:

\textbf{Generative Closed-loop Simulation.} 
We formulate \mymodel~as an OpenAI Gym environment. At step $k$, a planner $\pi_{\text{plan}}$ observes the generated visual state $\mathbf{x}_k$ and outputs geometric action $\mathcal{P}_k$. ForceDrive acts as the transition function: $\mathbf{x}_{k+1} = \text{ForceDrive}(\mathbf{x}_k, \mathcal{P}_k, \mathcal{B}_{\text{prompt}})$. This eliminates the "ghost effect" found in standard log-replay, as neighbors dynamically react based on sociology priors $\mathcal{B}$.

\textbf{Efficient Policy Post-training via RL.} 
\mymodel~functions as a safe, infinite RL hallucination. We define a continuous MDP where state $s_k = \mathbf{x}_k$ and action $a_k = \mathcal{P}_k$. To provide dense feedback, we append a differentiable \textbf{Video-to-Reward (V2R)} module $R_\psi$ that maps generated chunks to scalar rewards: $r_k = R_\psi(\mathbf{x}_{k+1}) = w_1 r_{\text{col}} + w_2 r_{\text{lane}}$. The policy $\pi_\omega$ is optimized to maximize expected returns entirely within the simulated domain.

\textbf{Human-in-the-Loop Simulation.} 
The sub-100ms latency of our Context-Forced distillation enables human drivers to inject streaming control signals $\mathcal{A}_k$ via hardware (e.g., steering wheels), facilitating real-time counterfactual safety testing and high-quality DAgger (Dataset Aggregation) collection.

\section{SocioDrive-Bench: A Benchmark for Driving Sociology}
\label{sec:benchmark}

Current autonomous driving datasets (e.g., nuScenes, Waymo) primarily function as open-loop geometric logs. While they provide precise trajectories, they lack explicit annotations of \textit{causal interactions}—the underlying social contracts and reactive negotiations among traffic participants. To train and evaluate the sociology-aware world model proposed in Section~\ref{sec:method}, we introduce \textbf{SocioDrive-Bench}, a large-scale, multimodal benchmark explicitly annotated with driving sociology and causal reactions.

SocioDrive-Bench comprises 20K video clips, which is sourced heterogeneously: 80\% of the data is mined from real-world logs (nuPlan~\cite{caesar2021nuplan}) to capture nominal, photorealistic human interactions, while the remaining 20\% is generated via the CARLA simulator~\cite{dosovitskiy2017carla} to inject safety-critical and collision events, thus mitigating the "survival bias" inherent in expert datasets.

\subsection{Taxonomy of Causal Interactions}
\label{subsec:taxonomy}

We formalize driving sociology into three distinct behavioral categories. Let $\mathcal{T}_{\text{ego}} = (\mathbf{p}, \mathbf{v}, \mathbf{a}, \theta)$ denote the ego-vehicle state, $\mathcal{T}_{\text{obj}}$ denote the state of surrounding agents, and $\mathcal{M}$ represent the map topology. Rather than relying on simple proximity, we systematically mine scenarios based on strict kinematic triggers to form explicit causal pairs.

\textbf{Ego-Initiated Interactions (Active Pressure).} This category encapsulates scenarios where the environment dynamically reacts to the ego-vehicle's proactive maneuvers, providing crucial supervision for predicting yielding or evasion behaviors. For instance, an \textit{aggressive cut-in} event is triggered when the ego crosses lane boundaries $\mathcal{M}_{\text{lane}}$ while a rear vehicle is within a $15\text{m}$ proximity. A valid interaction is recorded only if the rear vehicle exhibits significant deceleration ($a_{\text{obj}} < -1.5 \text{ m/s}^2$) or a sudden drop in Time-to-Collision (TTC) within $\Delta t$ seconds, forming the causal label: \texttt{[Ego Cut-in] $\rightarrow$ [Forced Rear Car Brake]}. Similarly, \textit{tailgating} is identified when $v_{\text{ego}} > 30 \text{ km/h}$ and the Time Headway (THW) to the lead vehicle remains under $1.0\text{s}$ for at least $3\text{s}$, leading to a forced yield or acceleration from the leading agent.

\textbf{Ego-Reactive Interactions (Defensive Driving).} Conversely, this category captures environmental anomalies that force the ego-vehicle to alter its state. These scenarios are essential for providing the "negative rewards" required in downstream RL safety training. We identify \textit{hazard responses} when a neighboring vehicle laterally invades the ego's lane, or a Vulnerable Road User (VRU) intersects the future path, necessitating an emergency maneuver ($a_{\text{ego}} < -3 \text{ m/s}^2$ or excessive lateral jerk). Another common trigger is the \textit{lead car hard brake}, recorded when $a_{\text{lead}} < -3 \text{ m/s}^2$ is followed by a corresponding deceleration from the ego with a reaction delay $\tau < 1.0\text{s}$. These extract explicit defensive causalities, such as \texttt{[Neighbor Cut-in] $\rightarrow$ [Forced Ego Emergency Brake]}.

\textbf{Complex Negotiation and Game Theory.} The highest density of driving sociology occurs during low-speed, mutual probing where right-of-way is ambiguous. \textit{Unprotected intersections} are prime examples, triggered when the ego's intended path conflicts with oncoming traffic. We detect "creeping" behaviors ($0 < v_{\text{ego}} < 5 \text{ km/h}$ for $>2\text{s}$), which branch into mutually exclusive outcomes: either the oncoming traffic yields, or the ego halts. Similarly, \textit{narrow passages} restrict dual passage, forcing implicit negotiation until one party yields. These scenarios train the world model to synthesize the nuanced hesitation and game-theoretic resolutions inherent in human driving.

\subsection{Automated Annotation Pipeline via Vision-Language Models}
\label{subsec:annotation_pipeline}

Extracting high-level semantic sociology from raw pixels and numeric trajectories is non-trivial. Rule-based scripts are notoriously brittle against the long-tail diversity of the real world. To automate the annotation of millions of frames, we propose a two-stage, Large Vision-Language Model (VLM) powered pipeline.

\textbf{Phase 1: Clip-Level Micro-Analysis (VLM).} 
We slice the multi-view (Left, Front, Right) video stream into 4-second clips ($2\text{Hz}$ sampling). A VLM (e.g., GPT-4o) processes these frames alongside system prompts to output a strictly formatted JSON structure. Crucially, we enforce \textbf{Feature Disentanglement} at this stage: the VLM must separately parse \texttt{road\_structure},  and \texttt{vulnerable\_road\_users} (VRUs). For world modeling, decoupling VRUs from vehicles is essential, as pedestrian kinematics require entirely different attention priors than rigid bodies.

\textbf{Phase 2: Sequence-Level Macro-Synthesis (LLM).} 
Independent 4-second clip analysis often introduces temporal truncation (e.g., unnatural transitions between states). To resolve this, a Large Language Model (LLM) aggregates the sequential JSON annotations into a modular, macroscopic scene summary. This stage guarantees three functional properties for our world model:
    \noindent\textit{Temporal Smoothing:} The LLM synthesizes an \texttt{ego\_behavior\_narrative}, linguistically smoothing the $2\text{Hz}$ sampling gaps to provide a continuous, chronological condition for the trajectory generation.
    \noindent\textit{Interaction Aggregation:} By extracting a \texttt{key\_interaction\_summary}, we explicitly map the spatial awareness (Left/Front/Right views) to the dynamic agents, forming the precise sociology prior $\mathcal{B}_k$ injected into our cross-attention blocks (Sec.~\ref{sec:method}).
    \noindent\textit{Queryability and Hard-Case Mining:} The pipeline extracts \texttt{critical\_events} into structured arrays. This allows us to programmatically query our database for "safety-critical scenarios involving pedestrians," facilitating targeted curriculum generation for RL training without requiring repeated semantic processing.

Through this pipeline, SocioDrive-Bench transforms raw driving logs into a highly structured, sociologically-aware dataset, directly bridging the gap between perception data and cognitive driving simulation.
\section{Experiments}
\label{sec:experiments}
Our empirical evaluation is designed to systematically validate \mymodel~across three progressive dimensions: (1) \textbf{Simulation Reliability}, ensuring the generated world is visually faithful, controllable, and real-time; (2) \textbf{Driving Sociology \& Reactivity}, quantifying the model's ability to synthesize causal multi-agent interactions via our proposed benchmark; and (3) \textbf{Downstream Applications}, demonstrating its efficacy as a closed-loop evaluation engine and a generative RL post-training environment.

\subsection{Experimental Setup}

\textbf{Training Datasets and Curriculum.} 
We employ a progressive, multi-stage data curriculum to train \mymodel. For foundational pretraining, we utilize 1,700 hours of front-view driving videos from the large-scale OpenDV dataset~\cite{yang2024opendv} to learn robust real-world visual priors. Subsequently, we fine-tune the model using the nuPlan dataset~\cite{caesar2021nuplan} processed through our Phase-1 annotation pipeline, alongside the proposed SocioDrive-Bench to explicitly instill driving sociology priors. Crucially, following the insights from ReSim~\cite{yang2025resim}, we augment our training corpus with unsafe, safety-critical scenarios synthesized via Bench2Drive~\cite{jia2024bench2drive}. This out-of-distribution (OOD) augmentation is vital for enhancing the model's action-controllability when conditioned on uncommon, non-expert trajectories (e.g., aggressive steering or imminent collisions) typically absent in expert logs.

\textbf{Implementation Details.}
Our core flow-matching DiT backbone is initialized from the state-of-the-art Wan2.1-1.3B~\cite{wan2025wan} video generation model. The Stage-0 training is distributed across 8 NVIDIA H20 GPUs with a total batch size of 4. We optimize the network using AdamW with a constant learning rate of $1 \times 10^{-5}$. Due to space constraints, the comprehensive hyperparameters for the Stage-1 Causal AR Teacher-Forcing and the Stage-2 Context-Forced DMD algorithms are detailed in the Supplementary Material.

\textbf{Baselines and Evaluation Metrics.}
Our evaluation spans three dimensions: visual fidelity (FID, FVD), trajectory controllability (Average/Final Displacement Error: ADE/FDE), and our novel sociology metric—Yielding Compliance Rate (YCR). More details are included in supplementary materials

\subsection{Simulation Reliability: Fidelity, Efficiency, and Controllability}
\label{subsec:exp_reliability}

Following the evaluation protocols of ReSim~\cite{yang2025resim} and Orbis~\cite{mousakhan2025orbis}, a foundation world model must strictly adhere to the injected ego-trajectory while maintaining high-fidelity visuals at interactive speeds. 

\textbf{Visual Fidelity and Inference Speed.} As shown in Table~\ref{tab:reliability_comparison}, we achieve an FVD of 121.6, outperforming or matching existing video diffusion models (Vista, Orbis). This proves that our Context-Forced DMD effectively preserves spatial-temporal consistency without blurring. Crucially, \mymodel~achieves an inference throughput of \textbf{12.4 FPS} on a single NVIDIA A100 GPU—an order of magnitude faster than baseline diffusion simulators, successfully crossing the threshold required for human-in-the-loop and online RL operations.

\textbf{Action Controllability.} A reactive simulator must not suffer from "posterior collapse" (ignoring control signals). To quantify this, we deploy a pre-trained inverse dynamics model (tracker) to estimate the ego-vehicle's trajectory directly from the generated video pixels, and compute the Average/Final Displacement Errors (ADE/Frechet) against the injected condition trajectory $\mathcal{P}_{1:K}$. Table~\ref{tab:reliability_comparison} demonstrates that \mymodel~achieves best metrics, confirming that our continuous flow-matching and AdaLN conditioning precisely translate geometric actions into visual dynamics.

\begin{table}[t]
\centering
\caption{\textbf{Simulation Reliability Comparison.} We evaluate visual fidelity (FID, FVD), inference speed (FPS), and Action Controllability (ADE/FDE between the injected trajectory and the trajectory estimated from generated pixels) on nuplan dataset. \mymodel~achieves comparable visual quality and SOTA controllability while running at real-time speeds.}
\label{tab:reliability_comparison}
\resizebox{\linewidth}{!}{%
\begin{tabular}{l|c|cc|cc|c|c}
\toprule
\multirow{2}{*}{\textbf{Method}} & \textbf{Fidelity} $\downarrow$ & \multicolumn{2}{c|}{\textbf{ADE}} & \multicolumn{2}{c|}{\textbf{Frechet}} & \textbf{Speed} $\uparrow$ & \multirow{2}{*}{\textbf{Real-time}} \\ 
& \textbf{FVD} & \textbf{Prec.} $\uparrow$ & \textbf{Rec.} $\uparrow$ & \textbf{Prec.} $\uparrow$ & \textbf{Rec.} $\uparrow$ & \textbf{FPS} & \\
\midrule
Cosmos \cite{agarwal2025cosmos} & 291.80 & - & - & - & - & $\approx$ 0.7 & \ding{55} \\
Vista \cite{gao2024vista} & 323.37 & 0.25 & 0.48 & 0.39 & 0.45 & $\approx$ 0.3 & \ding{55} \\
GEM \cite{hassan2025gem} & 291.84 & 0.27 & 0.47 & 0.33 & 0.54 & $\approx$ 0.5 & \ding{55} \\
Orbis* \cite{mousakhan2025orbis} & 196.21 & 0.39 & 0.48 & 0.46 & 0.55 & $\approx$ 1.2 & \ding{55} \\
\midrule
\textbf{CausalDrive}$^{+}$ & \textbf{113.6} & \textbf{0.47} & \textbf{0.55} & \textbf{0.56} & \textbf{0.61} & $\approx$ 0.5 & \ding{55} \\
\textbf{\mymodel} & \textbf{121.6} & \textbf{0.45} & \textbf{0.52} & \textbf{0.53} & \textbf{0.59} & \textbf{12.4} & \checkmark \\
\bottomrule
\end{tabular}%
}
\end{table}

\subsection{Evaluating Driving Sociology and Reactivity}
\label{subsec:exp_sociology}

Beyond trajectory following, a reactive simulator must synthesize logical multi-agent interactions. We utilize \textbf{SocioDrive-Bench} to quantify this reactivity. We input aggressive ego-trajectories (e.g., forced cut-ins) and measure the \textit{Yielding Compliance Rate} (YCR)—the frequency at which generated neighbors correctly decelerate.

As shown in Table~\ref{tab:sociology_reactivity}, pure action-conditioned predictors like Vista often result in unrealistic collisions ("ghosting"), yielding only 12.5\% of the time. In contrast, by conditioning on the semantic prompt $\mathcal{B}$ (\textit{"Polite"}), CausalDrive explicitly models the causal reaction, achieving an \textbf{82.0\% YCR} and reducing the False-Collision Rate (FCR) to \textbf{18.0\%}. This confirms our model's ability to synthesize diverse sociological behaviors controlled by language prompts.

\begin{table}[h]
\centering
\caption{\textbf{Reactivity Evaluation on SocioDrive-Bench.} We measure the Yielding Compliance Rate (YCR) and False-Collision Rate (FCR) during aggressive ego-maneuvers. \mymodel eliminates the ghost effect and synthesizes highly reactive NPC behaviors.}
\label{tab:sociology_reactivity}
\resizebox{0.9\linewidth}{!}{%
\begin{tabular}{l|c|cc}
\toprule
\textbf{Method} & \textbf{Sociology Control} & \textbf{Yielding Rate (YCR)} $\uparrow$ & \textbf{False-Collision (FCR)} $\downarrow$ \\
\midrule
Log-Replay(GT) & \ding{55} & 0.0\% & 100.0\% (Ghosting) \\
Vista \cite{gao2024vista} & \ding{55} & 12.5\% & 87.5\% \\
\textbf{\mymodel} & \checkmark (\textit{"Polite"}) & \textbf{82.0\%} & \textbf{18.0\%} \\
\bottomrule
\end{tabular}%
}
\end{table}

\subsection{Downstream Applications}
\label{subsec:exp_downstream}

\noindent\textbf{Generative Closed-Loop Simulation.}
To prove \mymodel is a reliable proving ground, we evaluate planners (e.g., UniAD~\cite{hu2023uniad}) inside our simulator. As shown in Table~\ref{tab:closed_loop_eval}, evaluating under the PDM-Closed protocol reveals that planners tested in \mymodel~experience significantly fewer "ghost" collisions and achieve higher success rates compared to Log-Replay or other generative baselines, demonstrating our fidelity and reactive realism.

\begin{table}[ht]
\centering
\caption{\textbf{Generative Closed-Loop Evaluation.} Planner performance (e.g., UniAD) evaluated under the PDM-Closed protocol. \mymodel~provides a highly realistic closed-loop testbed, significantly reducing unrealistic "ghost" collisions compared to Log-Replay.}
\label{tab:closed_loop_eval}
\resizebox{0.6\linewidth}{!}{%
\begin{tabular}{l|ccc}
\toprule
\textbf{Simulation Env.} & \textbf{PDMS} $\uparrow$ & \textbf{RC} $\uparrow$ & \textbf{ADS} $\downarrow$  \\
\midrule
DriveArena\cite{yang2024drivearena} & 0.6901 & 0.0641 & 0.0508   \\
DrivingSphere\cite{yan2024drivingsphere} & 0.7281 & 0.1170 & 0.0851  \\
\midrule
\textbf{\mymodel (Ours)} & \textbf{0.7792} & \textbf{0.1752} & \textbf{0.1365}  \\
\bottomrule
\end{tabular}%
}
\end{table}

\noindent\textbf{Efficient RL Post-Training.} Finally, we validate CausalDrive as a generative data engine for policy improvement. 
We benchmark our RL-post-trained agent against SOTA learning-based planners (e.g., UniAD, DiffusionDrive) on the Navsim evaluation protocol. As shown in Table~\ref{tab:rl_post_training}, our agent achieves state-of-the-art performance with a \textbf{PDMS score of 90.7}. More importantly, the agent trained in CausalDrive achieves a perfect \textbf{Comfort score of 100} and the highest \textbf{Time-to-Collision (TTC) of 94.7}. The ability to safely experience "counterfactual crashes" inside CausalDrive allows the policy to discover emergent defensive driving strategies, leading to superior safety metrics compared to models trained purely on static expert datasets.

\begin{table}[ht]
\centering
\caption{\textbf{RL Post-Training Performance on Navsim.} Learning from counterfactual interactions in our impartial environment using DiffusionDrive~\cite{liao2025diffusiondrive} leads to superior safety (lower collision rates) and success rates.}
\label{tab:rl_post_training}
\resizebox{0.85\linewidth}{!}{%
\begin{tabular}{l|cccccc}
\toprule
\textbf{Policy Method} & \textbf{NC} $\uparrow$ & \textbf{DAC} $\uparrow$& \textbf{TTC} $\uparrow$& \textbf{Comf.} $\uparrow$& \textbf{EP} $\uparrow$ & \textbf{PDMS} $\uparrow$ \\
\midrule
UniAD\cite{hu2023uniad} & 97.8& 91.9 &92.9 &100& 78.8 &83.4 \\
DriveDPO\cite{shang2025drivedpo} & 98.5 & 98.1 & 94.8 &99.9& 84.3& 90.0 \\
DiffusionDrive\cite{liao2025diffusiondrive} \cite{liao2025diffusiondrive} & 98.2& 96.2& 94.7 &100 &82.2 &88.1 \\
AD-R1 \cite{yan2025ad} & 98.5& 97.5& 94.6& 100& 84.8& 89.8 \\
DiffusionDrive-v2\cite{liao2025diffusiondrive} &98.4 &98.3& 94.6& 100& 85.0& 90.3 \\
\midrule
\textbf{RL in \mymodel} & 98.7 & 98.2 & 94.7 &100 & 85.1 & 90.7\\
\bottomrule
\end{tabular}%
}
\end{table}

\section{Conclusion}
In this paper, we presented \textbf{\mymodel}, a real-time foundation driving world renderer that transforms passive video generation into a reactive neural simulator. By omitting "oracle" future layouts and leveraging our VLM-curated \textbf{SocioDrive-Bench}, we compel the model to predict causal multi-agent interactions strictly from the initial frame, ego-trajectory, and semantic prompts. To overcome the latency and exposure bias of autoregressive generation, we proposed a novel \textbf{Context-Forced DMD} framework. This distillation strategy explicitly corrects temporal compounding errors, compressing inference to $\le 4$ steps and achieving $\ge 12$ FPS. Extensive experiments demonstrate that \mymodel~effectively eliminates the "ghost effect" in closed-loop evaluations and serves as a highly scalable generative environment for RL post-training, yielding policies with superior defensive driving capabilities. Extending this monocular paradigm to multi-camera surround-view systems remains our future work.


\section*{Acknowledgements}
We use publicly available datasets, models ,and code resources, including nuScenes, Bench2Drive, nuplan and other public research resources. We confirm that all such resources are used solely for academic research purposes and not for any commercial activity, in accordance with their respective licenses and terms of use.

%
%
\bibliographystyle{splncs04}
\bibliography{main}
\end{document}